\pdfoutput=1

\documentclass[11pt]{article}

\usepackage{emnlp2021}

\usepackage{times}
\usepackage{latexsym}

\usepackage{CJKutf8}
\usepackage{amsmath}
\usepackage{amssymb}

\usepackage{array}
\usepackage{booktabs}

\usepackage{graphicx}
\usepackage{tikz}
\usepackage{tikz-dependency}

\newcommand{\ourmodels}{SERA}
\newcommand{\tabincell}[2]{\begin{tabular}{@{}#1@{}}#2\end{tabular}}

\usepackage[T1]{fontenc}

\usepackage[utf8]{inputenc}

\usepackage{microtype}

\title{Entity Relation Extraction as Dependency Parsing in Visually Rich Documents}

\author{Yue Zhang,  Bo Zhang, Rui Wang, Junjie Cao, Chen Li, Zuyi Bao  \\
Alibaba Group, China  \\
\{\tt shiyu.zy, klayzhang.zb, masi.wr, puji.lc, \\ 
junjie.junjiecao, zuyi.bzy \}@alibaba-inc.com}

\author{
Yue Zhang$^1$,Bo Zhang$^1$, Rui Wang$^2$\thanks{$~$ Corresponding author. The author's contributions were carried out while at Alibaba Group. His current affiliation is Vipshop (China) Co., Ltd.}, Junjie Cao$^{1}$, Chen Li$^1$, Zuyi Bao$^1$ \\
Alibaba Group, China \\
{\tt $^1$\{shiyu.zy,klayzhang.zb,puji.lc,junjie.junjiecao,zuyi.bzy\}} \\ 
{\tt @alibaba-inc.com} \\
{\tt $^2$mars198356@hotmail.com}
}

\begin{document}
\maketitle
\begin{abstract}
Previous works on key information extraction from visually rich documents (VRDs) mainly focus on labeling the text within each bounding box (i.e., \emph{semantic entity}), while the relations in-between are largely unexplored. In this paper, we adapt the popular dependency parsing model, the biaffine parser, to this \emph{entity relation extraction} task. Being different from the original dependency parsing model which recognizes dependency relations between words, we identify relations between groups of words with layout information instead. We have compared different representations of the semantic entity, different VRD encoders, and different relation decoders. For the model training, we explore multi-task learning to combine entity labeling and relation extraction tasks; and for the evaluation, we conduct experiments on different datasets with filtering and augmentation. The results demonstrate that our proposed model achieves 65.96\% F1 score on the FUNSD dataset. As for the real-world application, our model has been applied to the in-house customs data, achieving reliable performance in the production setting.
\end{abstract}

\section{Introduction}

In real-life scenarios, there are many types of visually rich documents (VRDs), such as invoices, questionnaire forms, declaration materials and so on.
These documents contain abundant layout information which helps us to understand the content while texts alone are not enough. 
In recent years, many works focus on how to extract key information from VRDs based on the results of OCR (Optical Character Recognition), which recognizes bounding boxes and texts within the boxes \cite{liu2019gcn_labeling,yu2020pick}.
Each bounding box contains 1) a group of words that belong together from a semantic and spatial standpoint and 2) visual features such as layout, tabular structure and font size of the boxes in the document.
We call such bounding boxes and texts within the boxes \emph{semantic entities}\footnote{In different papers, the bounding boxes are called differently, such as semantic entities, text segments, etc. In this paper, we follow the naming convention used in the paper of the FUNSD dataset \cite{funsd_data}.}, and each entity contains the word group and layout coordinates\footnote{Other visual features such as font size, colors and so on are not provided in the FUNSD dataset, thus not considered in this work.}.

\begin{figure*}
  \centering
  \includegraphics[width=.9\textwidth]{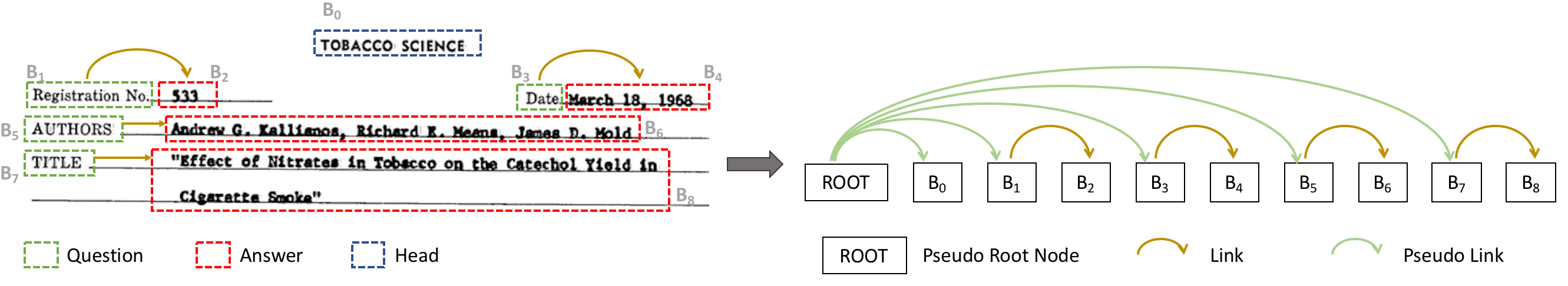} 
  \caption{Left part shows one visually rich document from the FUNSD data. Group of words with one bounding box means one semantic entity and we number each entity as $B_i$.
  Different colors of boxes mean their different entity labels as legends under the example list. 
  Relation links between semantic entities always point from key entities to value entities. 
  We convert entity relations in the VRD to one tree and add the pseudo root node following the similar setting in dependency tree, and then link zero-head entities to the pseudo node as the pseudo links shown in the right part.}  
  \label{funsd_example}
\end{figure*}

Key information extraction (KIE) is such a task to analyze visually rich documents, which usually contains two steps, \emph{entity labeling} and \emph{entity relation extraction}. Similar to named entity recognition (NER) and relation extraction in the traditional natural language processing (NLP), \emph{entity labeling} aims to assign predefined labels to the semantic entities in VRDs \cite{funsd_data, liu2019gcn_labeling, yu2020pick}, and \emph{entity relation extraction}\footnote{ To avoid confusion with the entity linking in knowledge graphs, we name the task as entity relation extraction instead of entity linking used in \citet{funsd_data}, as this task aims to discover relations between semantic entities} predicts relations between these semantic entities. Compared with NER and relation extraction in the traditional NLP, KIE from VRDs is a more challenging task. First, a normal (named-) entity in plain text does not contain layout information as those semantic entities in VRDs. Second, normal relation extraction predicts the relation between two \textit{given} mentions, while relation extraction in VRDs needs to predict the relation between \textit{any} two semantic entities in the document. As Figure~\ref{funsd_example} (left) illustrates, the entity labeling task is to tag ``533'' with the label ``Answer'', ``Registration No.'' with label ``Question''. However, which question can be answered by ``533'' remains unknown without \emph{entity relation extraction}. Compared with labeling, the entity relation extraction task is less explored, but its benefits at least include: 1) providing additional structural information closer to human comprehension of the VRDs, and 2) being easier to be transferred to other domains when the predefined label set changes. Therefore, in this paper, we concentrate on the task of semantic entity relation extraction which discovers the relation between two groups of words with layout information, as the yellow links in Figure~\ref{funsd_example} (left) show.

As another similar task to entity relation extraction in VRDs, dependency parsing aims to find out syntactic relations between words of an input sentence, which has been widely studied for decades. 
Both of these two tasks capture pairwise relationship between basic units of the input data.
We adapt the popular biaffine dependency parser which utilizes the biaffine attention to compute scores between words \cite{dozat2016deep} into the entity relation extraction task due to their similarity.
Since visual features play an important role in the VRDs, we introduce abundant layout information into different layers of the model to enhance the original text-only biaffine model: 

\begin{itemize}
    \item At the \textbf{entity representation} layer, we use the LayoutLM \cite{xu2020layoutlm} to encode both the word group and coordinates.
    \item At the \textbf{document encoder} layer, we utilize graph convolutional networks (GCN) to combine textual and visual information in VRDs by mapping layout information into graph edge representation between entities \cite{liu2019gcn_labeling, yu2020pick}.
    \item At the \textbf{relation scorer} layer, we extract relative position features between entities according to their coordinates.
\end{itemize}

Apart from the above, inspired by the joint POS tagging and dependency parsing model \cite{pos_dep}, we propose the multi-task learning for both entity labeling and relation extraction to further improve the performance.

Abundant detailed experiments are conducted to verify our approach of applying the biaffine dependency parser to semantic entity relation extraction task in VRDs. Our proposed relation extraction model achieves 65.96\% F1 score on the FUNSD dataset, demonstrating the effectiveness of our model. As for the real-world application scenario, our model has also been applied to the in-house customs data, achieving reliable performance in the production setting.

The contributions of this paper are as follows:
\begin{itemize}
\item We adapt the biaffine model used in dependency parsing to the entity relation extraction task and achieve 65.96\% F1 score on the FUNSD dataset.
\item We conduct detailed experiments to compare different representations of the semantic entity, different VRD encoders, and different relation decoders to better understand this task.
\item We apply our model to the real-world customs data with different layouts and achieve high performance in the production setting.
\end{itemize}

\begin{figure*}
  \centering
  \includegraphics[width=.9\textwidth]{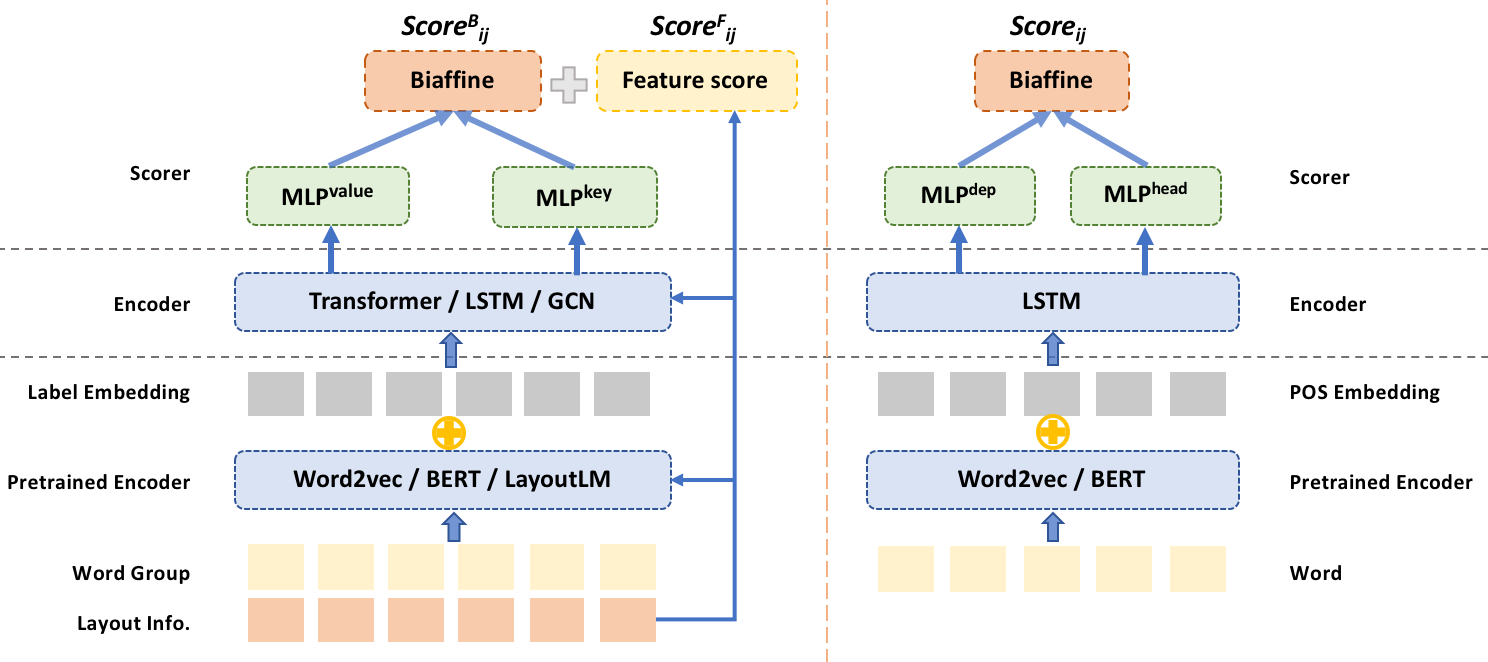} 
  \caption{The architecture of our proposed entity relation extraction model (left) and the biaffine parser model (right).}  
  \label{modelimg}
\end{figure*}

\section{Related Work}
Visually rich document understanding includes many tasks, such as layout recognization \cite{publaynet,li2020docbank}, table detection and recognition \cite{tablebank, zhong2019PubTabNet} and key information extraction \cite{gralinski2020kleister, guo2019eaten, sroie, funsd_data, majumder2020representation}.
Our paper focuses on the key information extraction task which contains two subtasks, entity labeling and relation extraction. The former subtask tags entities with predefined labels, such as Task 3 on the SROIE data released by \citet{sroie}, while the latter discovers relations between entities, such as Subtask C(3) on the FUNSD data \cite{funsd_data}.

To encode the semantic entity in VRDs, \citet{yu2020pick} and \citet{kie_feature} replace BiLSTM (Bi-directional Long Short-Term Memory) used by \citet{liu2019gcn_labeling} with BERT \cite{devlin2018bert} or RoBERTa \cite{roberta}.
\citet{xu2020layoutlm} propose LayoutLM, which adds the 2-D position embedding into language model based on BERT and pretrain their language model on large-scale scanned document images with more visually-related loss function.
Experiments verify that encoding the word group and layout coordinates at the same time is more effective for VRD understanding.
LayoutLMv2 additionally introduces visual embedding into input layer and integrates a spatial-aware self-attention mechanism into the Transformer architecture \cite{xu2020layoutlmv2}.
And LayoutLMv2 performs better than LayoutLM in downstream VRD understanding tasks.

While encoding VRDs, previous works take entity labeling task as sequence labeling and re-implement the named entity recognition (NER) framework \cite{lample2016neural} but ignore layout information.
Then, many works introduce a GCN-based module to encode layout information and combine textual and visual information together \cite{liu2019gcn_labeling, yu2020pick, kie_feature, carbonell2021named}.
In the GCN module, \citet{liu2019gcn_labeling, yu2020pick} take layout features between entity $b_i$ and $b_j$ as edge embedding to update entity representation while \citet{kie_feature} prune irrelevant nodes in graph according to same x-axis or y-axis coordinates to get the adjacency matrix.

To predict relations between entities, \citet{funsd_data} provide one simple approach which concatenates the representations of two entities and use multi-layer perceptron (MLP) to obtain the relation score between entities. 
\citet{carbonell2021named} also use the MLP scorer but take GNN as document encoder instead of BERT and perform better.
In the field of dependency parsing, \citet{dozat2016deep} propose the biaffine attention mechanism to compute scores between words, and achieve better performance than the MLP mechanism used by \citet{dep_parser_mlp}. As the biaffine attention is widely used in other tasks like NER \cite{yu2020-ner_as_dep} and semantic role labeling \cite{li2019biaffinesrl}, we propose to use it for the entity relation extraction task in this work.

\section{Entity Relation Extraction as Dependency Parsing}
\label{model_decl}
Both semantic entity relation extraction and dependency parsing tasks aim to decide whether there exists relation between two entities/words and assume that links always point from key/head unit to value/modifier unit shown in Figure~\ref{funsd_example}.
Therefore, we can draw lessons from the dependency parsing exploration as it has been studied for several decades and achieved great progress.
Biaffine parser, a strong model in dependency parsing, achieves competitive performance and is widely used in different scenes and tasks.
This section introduces how to apply the biaffine parser to our relation extraction task according to their similarity and difference.

\subsection{Task Definition} 
Each scanned visually rich document is composed of a list of semantic entities, and each entity composes of a group of words and coordinates of the bounding box, defined as $b_i = \{ [w_i^1,…,w_i^m], [x_i^1, x_i^2, y_i^1, y_i^2]\}$, where $[w_i^1,…,w_i^m]$ mean the word group, $x_i^1$/$x_i^2$ and $y_i^1$/$y_i^2$ are left/right x-coordinates and top/down y-coordinates respectively. 
Documents in our used dataset are annotated with the label of each entity and relations between entities.
We represent each annotated document as $D = \{[b_1, ... b_n], [l_1, ..., l_n], [(b_1, b_{h_1}), ..., (b_m, b_{h_m})]\}$, where
$l \in L $ is the label of each entity and $L$ is the predefined entity label set.
$(b_i, b_{h_i})$ mean the relation between entities $b_i$ and $b_{h_i}$, and the link points from $b_{h_i}$ to $b_i$.
Notably, the entity may exist relations with more than one entity or does not have relation with any other entities. 

\subsection{Biaffine Parser}
As Figure~\ref{modelimg} (right) shows, biaffine dependency parser takes word and POS-tag embedding as the word representation, and uses multi-layer BiLSTM to encode the input sentence.
Then, two MLP modules are used to strip away information not relevant to the current link decision.
At last, the biaffine attention mechanism is utilized to compute the score of the dependency link between words.

We explore various aspects of applying the biaffine parser to our relation extraction in VRDs due to their similarity.
Especially, we take the layout information into consideration besides the text itself, compared to the regular dependency parsing.
In our proposed entity relation extraction model, we exploit important layout information at different processing levels, including entity encoder, document encoder and relation scorer, as Figure \ref{modelimg} (left) shows.
We name our proposed model as \textsc{\ourmodels{}} (\textbf{S}emantic \textbf{E}ntity \textbf{R}elation extraction \textbf{A}s dependency parsing).
Details of our proposed \textsc{\ourmodels{}} are discussed in the following subsections.

\subsection{Entity Representation} 
\label{node_rep_section}

At the input layer, in order to obtain better entity representations, we compare different ways to encode the information of semantic entity, containing the word group and the layout features. 
In this work, we take the advantage of widely used pretrained models, including context-free word vector from word2vec \cite{mikolov2013word2vec}, contextualized representations from BERT and LayoutLM. Specially, LayoutLM  introduce coordinate information from bounding boxes during pretraining which is very suitable for our scenario.

In addition, we make use of the label of each semantic entity, such as ``Question'', ``Answer'' in FUNSD label set as Figure \ref{funsd_example} shows.
We map the entity labels into label embedding, as POS-tag embedding in dependency parsing.
Then, we concatenate the entity representation and label embedding as the input of the document encoder for each semantic entity, as the following equation shows:
\begin{equation}
\mathbf{e}_i = \mathbf{b}_i \oplus \mathbf{l}_i
\end{equation}
where $\mathbf{l}_i$ means entity label embedding, and $\mathbf{b}_i$ means the representation of semantic entity, which can be obtained from the above-mentioned three pretrained models, e.g. word2vec, BERT and LayoutLM .

\subsection{Document Encoder}

We compare different document encoders, including transformer, BiLSTM, and GCN, for better encoding the information of the semantic entities in VRDs. 
Specifically, we  feed the representation of entity into the document encoder and obtain the output of the encoder as the contextual representation of the entity. Details of BiLSTM and transformer can refer to \citet{lample2016neural} and \citet{vaswani2017attention}, respectively.

For GCN encoder, initial entity representation in graph is computed as subsection \ref{node_rep_section} shows.
While updating the representation of the entity and edge, we follow the computation of  \citet{liu2019gcn_labeling}.
The edge embedding consists of 2 layout features, as the following equation shows:
\begin{equation}
\mathbf{r}_{i, j} = [x_{i,j}, y_{i,j}]
\end{equation}
where $x_{ij}$ and $y_{ij}$ are horizontal and vertical distance between the two entity boxes respectively:
\begin{equation}
\begin{aligned}
\label{ra}
\scriptsize
&x_{i, j} = min(| x_i^1 - x_j^2 |, | x_j^1 - x_i^2 |) \\
&y_{i, j} = min(| y_i^1 - y_j^2 |, | y_j^1 - y_i^2 |)
\end{aligned}
\end{equation}

For entity $b_i$, we extract features $\mathbf{h}_{ij}$ of each neighbour $b_j$ by concatenating the representation of the two entities  and their corresponding edge.
\begin{equation}
\label{equation_h}
\mathbf{h}_{ij} = \mathbf{e}_i \oplus \mathbf{r}_{i,j} \oplus \mathbf{e}_j
\end{equation}

Then, we update the representation of entity and edge  so that each entity can extract relevant information from other entities according to the document layout information, as the following equation shows: 
\begin{equation}
\begin{aligned}
&\mathbf{e}_i^{'} = \sum_{j=1}^{n} \alpha_{ij}\mathbf{h}_{ij} \\
&\mathbf{r}_{i,j}^{'} =  \mathbf{W}^r \mathbf{h}_{ij} + \mathbf{b}^r
\end{aligned}
\end{equation}
and $\alpha_{ij}$ is the attention weight and can be computed as follows:
\begin{equation}
\alpha_{ij} = \frac {exp(LeakyRelu(\mathbf{W}^a \mathbf{h}_{ij}))} {\sum_{j=1}^{n}  exp(LeakyRelu(\mathbf{W}^a \mathbf{h}_{ij}))} 
\end{equation}
Where, $n$ means the number of entities in a document.

\subsection{Relation Scorer}

Following the biaffine parser, we firstly apply MLP module to drop trivial information which is unrelated to current relation decision.
Two MLP are used to generate the different representation of key and value roles in each relation link, which indicates the direction of arc in Figure~\ref{funsd_example}.
\begin{equation}
\begin{aligned}
\label{ra}
\scriptsize
\mathbf{h}_i^{key} = F(\mathbf{W}^{key} \mathbf{e}_i^{'} + \mathbf{b}^{key}) \\
\mathbf{h}_j^{value} = F(\mathbf{W}^{value} \mathbf{e}_j^{'} + \mathbf{b}^{value}) \\
\end{aligned}
\end{equation}
where $F$ is an activation function.

Then, we use biaffine attention to compute the score between two semantic entities as follows:
\begin{equation}
\begin{aligned}
\label{ra}
\scriptsize
Score_{i, j}^{B} = & \mathbf{h}_i^{key}  \mathbf{W}_1^B  \mathbf{h}_j^{value} +  \mathbf{h}_i^{key}   \mathbf{W}_2^B
\end{aligned}
\end{equation}
Such biaffine mechanism can capture pairwise relationship between entities better.

We also use layout information $\mathbf{r}_{i,j}$ as external features to help the model  predict relations between entities better.
Such layout features indicate the position relationship between entity $b_i$ and entity $b_j$: left-to-right or top-to-down.
Empirically, we observe that entities in the left-to-right or top-to-down order are more likely to exist relations.
We use MLP to compute the layout feature score as follows:
\begin{equation}
Score_{i, j}^F = \mathbf{W}^F  \mathbf{r}_{i,j} + \mathbf{b}^F
\end{equation}

Lastly, we add biaffine score with layout feature score together as the score of the relation between entity $b_i$ and entity $b_j$:
\begin{equation}
Score_{i, j} = Score_{i, j}^B + Score_{i, j}^F
\end{equation}

\subsection{Relation Decoder}
Based on relation scores between entities, two different relation decoding methods decide different loss functions of our training objective.

The first method is to judge whether there exists relation between any two entities in each VRD and such way is similar to semantic role labeling (SRL).
In such setting, we take the relation prediction as a binary classification task and use binary cross entropy loss as \citet{funsd_data}.

The second is to choose one head entity from all entities in one VRD for the current one, which is similar to the decoder in dependency parsing.
This method means that each entity must have exactly one head entity, namely single-head constraint.
Now, relation prediction is seen as a multi-classification task and use softmax cross entropy loss as \citet{dozat2016deep}.

\section{Experiment Settings}

\begin{table}
\small
\begin{center}
\begin{tabular}{p{1.2cm} | p{0.4cm} p{0.7cm} p{0.7cm} | p{0.4cm} p{0.7cm} p{0.7cm} }
\hline   &  \multicolumn{3}{c|}{\textbf{Train}}  &\multicolumn{3}{c}{\textbf{Test}}  \\ 
 & \#Docs & \#Entities & \#Rels &  \#Docs & \#Entities & \#Rels \\ \hline
\hline \textsc{FUNSD} & 149 & 7411 &3902&50 & 2332 &1048      \\
\ \  \textsc{w/ Aug} &  298  & 14,822  &   7804 &  50 & 2332 &  1048      \\
\hline \textsc{Customs} &    1308  &  93,627  &  16,146  &  329 & 23,365 & 4305     \\  \hline
\end{tabular}
\end{center}
\caption{Data statistics of the FUNSD and customs datasets.}
\label{data_statistic}
\end{table}

\subsection{Datasets}
We conduct experiments on the FUNSD\footnote{The FUNSD data can be downloaded from https://guillaumejaume.github.io/FUNSD/.} data, which is published by \citet{funsd_data} for the form understanding task. 
Moreover, to verify our proposed model, we also collect real-world dataset from the customs scenario. 

\textbf{FUNSD} is composed of 199 fully annotated, scanned forms with comprehensive annotations to address form understanding tasks including entity labeling and relation extraction. 
We follow the data split as \citet{funsd_data}, and detailed data statistics are listed in Table~\ref{data_statistic}, including the entity/relation distribution in FUNSD.

\begin{figure}
  \centering
  \includegraphics[width=.48\textwidth]{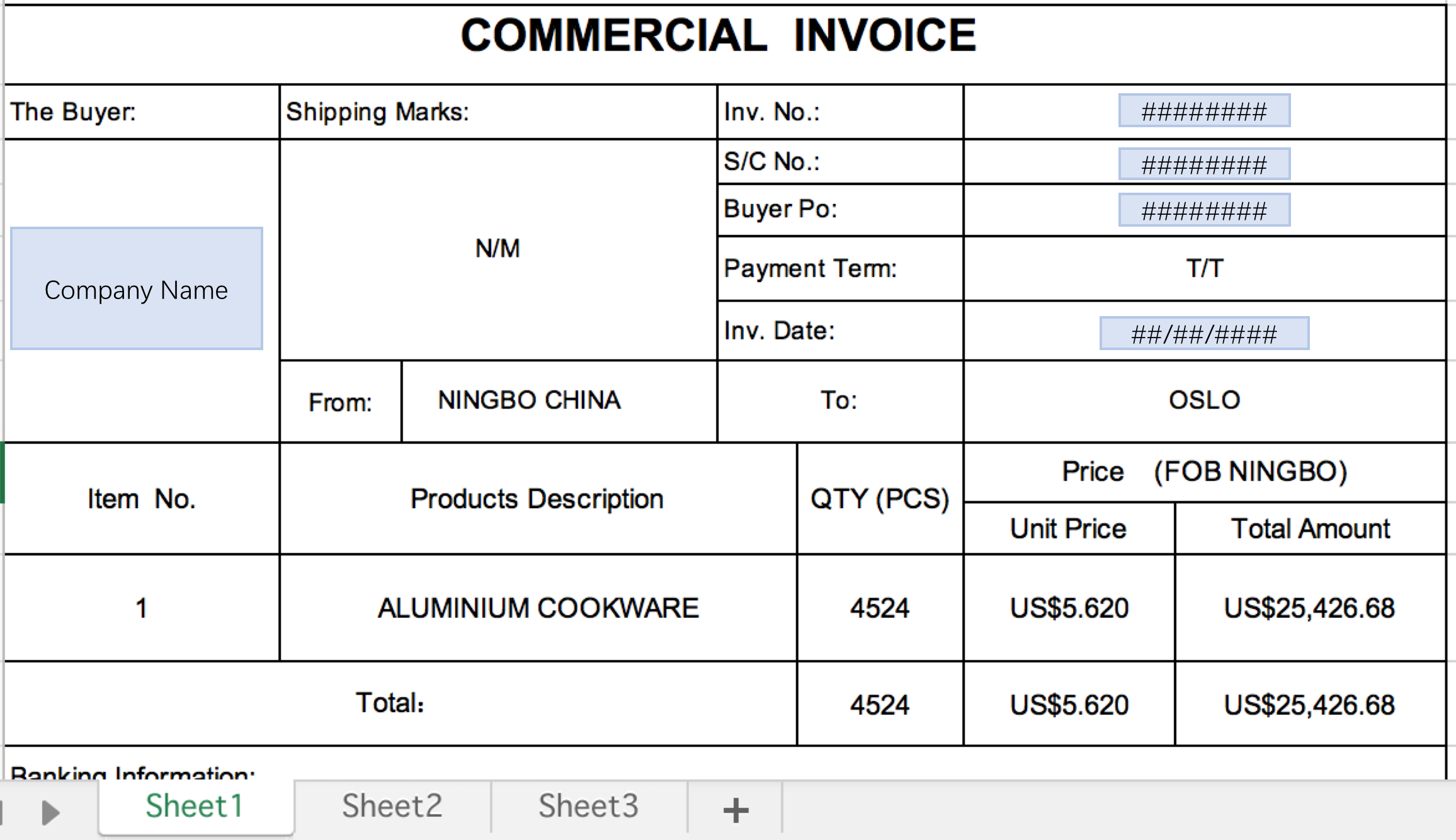} 
  \caption{One commercial invoice example in customs data and its filetype is Excel. Some sensitive information are replaced by blue blocks.} 
  \label{custom_invoice_example}
\end{figure}

\textbf{Customs Data} consists of about 1,600 customs declaration documents in different layouts and languages collected by us. There are four types of documents: packing list, invoice, sales contract and customs declaration form, and each kind of document provides different information  which is useful to apply to the customs.
Figure~\ref{custom_invoice_example} gives one invoice example, providing unit price, quantity and other  details.
Customs documents may be in Chinese or English, and their format may contain Word, Excel, PDF or image.
We parse these documents by a self-developed OCR tool to get semantic entities in each VRD.
We organize crowd-sourcing to annotate labels of entities given the predefined label set, containing 48 kinds of label which are important for customs information extraction system.
We can get the key entities according to the map dictionary from each predefined entity label to its all possible names in VRDs due to these names are enumerable.
Then, we link the entities from keys to values respectively with same labels.
We finally get the annotated customs data annotated with labels of entities and relations between entities. 
The scale of our collected customs data is much bigger than the FUNSD data as Table~\ref{data_statistic} shows.

\subsection{Data Preprocessing}

\textbf{Multi-Head \& Zero-Head Entities.} \label{multi-zero-head}
In terms of dependency parsing, one word must have one and only one head (the single-head constraint). 
However, zero-head entity that has no relations with any other entities or multi-head entity that has multiple heads does appear in our dataset. 
For zero-head entity, we add a pseudo root entity and link these zero-head entities to the pseudo entity as Figure~\ref{funsd_example} shows.
For multi-head entity, we randomly remain one head entity and delete others to get single-head entity while under single-head constraint. 
In FUNSD, there are 324 (/4,236) and 16 (/1,048) multi-head entities in train/test data respectively, accounts for a small part in all data. 

\textbf{Auto Labels.} 
It is intuitive that type-tagged entity will ease the prediction of relations between semantic entities.
We employ an effective entity labeling model consisting of two modules: entity encoder and label scorer.  
We take LayoutLM to encode the document and get the entity representation in a similar way as our relation extraction model. 
We also introduce three layout features $w_i, h_i, c_i$ into our entity representation and map the features into 10-dim embedding.
$w_i$ and $h_i$ mean the width and height of the bounding box and $c_i$ means the length of characters in word group of each semantic entity.
After concatenating the feature embedding and LayoutLM output, we pass them into the MLP scorer to compute the score of each candidate label.
By this way, we get the auto label of our relation extraction data\footnote{We train the labeling model on the whole training data and predict the auto labels of the test data. And we split training data into 5-fold, and train model with 4-fold to generate automatic labels of the left 1-fold training data.}.

\subsection{Evaluation Metrics}
To evaluate our semantic entity linking model, we take the entity-level precision, recall and F1 score as measure standard.
Under single-head constraint, we ignore the links pointed from pseudo root entity in gold and predicted results for fair comparison with other works.

\subsection{Parameters}
We investigate several pretrained language models to obtain entity representation, i.e, word2vec, BERT, and LayoutLM. 
For word2vec, we obtain the entity representation by averaging embeddings of the words contained in one entity; and for BERT/LayoutLM, we use the base model and take the hidden state output of the first subword of word group as the whole entity representation. 
Therefore, the representation dimension of words within bounding box is 100 while using word2vec and 768 while using BERT or LayoutLM.
We use 100-dim embedding to represent the entity labels, so our entity representation is 200 or 864\footnote{While using transformer, it's difficult to set the number of heads in multi-head self-attention if dim of entity representation is 868. Here, we use a 96-dim label embedding instead.}.

To encode the whole VRD, we investigate 1-layer BiLSTM or 1-layer transformer or 2-layer GCN encoders. The hidden state dimension of BiLSTM and transformer is 300 and the dimension of output edge and entity representations generated by GCN is 100.

The learning rate for BERT and LayoutLM is set to 1e-5 and others to 1e-2.
The model are trained for 50 iterations on FUNSD data and 100 iterations on customs data\footnote{We train the model for 50/100 iterations, and then predict test set on the trained model.}. And each iteration we traverse the whole training data under all settings.

\begin{table}
\begin{center}
\small
\begin{tabular}{ l | p{0.72cm}<{\centering} p{0.72cm}<{\centering} p{0.75cm}<{\centering}}
\hline
&\textbf{P} & \textbf{R} & \textbf{F1}  \\  \hline \hline
\multicolumn{4}{l}{\textbf{Previous works: our reimplementation}} \\
\hline \ \tabincell{l}{\textsc{Biaffine}\\ \citet{dozat2016deep}}   &  0.0746   & 0.1131 &   0.0899     \\
\hline \ \tabincell{l}{\textsc{Bert} \\ \citet{funsd_data} } & 0.2959  & 0.2872 & 0.2915 \\ 
\hline \ \tabincell{l}{\textsc{LayoutLM} \\ \citet{xu2020layoutlm} } &  0.3041 & 0.3082  & 0.3062  \\ 
\hline \ \tabincell{l}{\textsc{GNN + MLP} \\ \citet{carbonell2021named} } &  - & -  & 0.39  \\ 
\hline \hline

\multicolumn{4}{l}{\textbf{Our models}} \\ 

\hline \ \ \textsc{ - feature scorer} & 0.6049 & 0.6020 & 0.6035 \\ 
\hline \ \  \textsc{ - Auto Label} &   0.6046  &  0.6756  &  0.6381  \\ 
\hline \ \  \textsc{\ourmodels{}} &   0.6189  &  0.6756  &    0.6460     \\ 
\hline \ \  \textsc{ + Gold label} & 0.7033  & 0.7576 & 0.7294 \\
\hline \ \ \textsc{ + MTL} &  \textbf{0.6374}  &  0.6727  &  0.6546  \\
\hline \ \ \textsc{ + MTL,Aug}  &  0.6368   & \textbf{0.6842} & \textbf{0.6596}   \\  
\hline 

\end{tabular}
\end{center}
\caption{Performance of entity relation extraction task on the FUNSD test data of previous works and our model with different but important settings. We re-implement previous works after application to entity relation extraction task, except for the work of \citet{carbonell2021named}. We report their published experiment results in their paper. }
\label{full_exp_table}
\end{table}

\section{Experiment Results}

\subsection{Overall Results}

We adapt the biaffine model from the dependency parsing task to our entity relation extraction task, and conduct detailed experiments on FUNSD dataset. Experimental results are shown in Table~\ref{full_exp_table}.

\textbf{Previous works.} 
Firstly, we train the original biaffine model \cite{dozat2016deep} after replacing each word and its POS-tag with word group and entity label of each entity, but achieve poor performance.
Then, we re-implement the entity relation extraction model proposed by \citet{funsd_data}, which consists of BERT as the entity encoder and MLP as the relation scorer.
And our re-implement results are much higher than the performance reported in their paper (0.04\% F1).
We replace BERT with LayoutLM and keep other parts unchanged to encode the layout coordinate information into relation extraction task and model performance improves a little.
\citet{carbonell2021named} also utilize MLP link scorer but encode documents with k-layer GNN instead of BERT or LayoutLM, and they achieve higher performance than other previous works. 

\textbf{Our Models.}
We propose our semantic entity relation extraction model based on the architecture of the biaffine parser as Section~\ref{model_decl} describes.
We apply LayoutLM/GCN as our entity/document encoder and optimize our models with softmax cross entropy loss under the single-head constraint.
Results show that our proposed \textsc{\ourmodels{}} achieves much higher performance than previous works by a large margin. 
Performance improvement demonstrates that layout information plays an important role in entity relation extraction task.
Two ablation experiments verify the effectiveness of the layout feature scorer and auto labels of entities. 
Our \textsc{\ourmodels{}} model can further improve the performance with two training strategies: data augmentation and multi-task learning of entity labeling and relation extraction.

Detailed experiments about our different explorations for entity relation extraction task are discussed in the following subsections.

\begin{table}
\begin{center}
\small
\begin{tabular}{ l |  c   c  c }
\hline
 & \textbf{P} & \textbf{R} & \textbf{F1}  \\ \hline
\hline \textsc{word2vec} & 0.0187 & 0.0403 & 0.0256     \\
\hline \textsc{BERT} & 0.5539 & 0.5887 & 0.5708  \\  
\hline \textsc{LayoutLM} & \textbf{0.6189}  &  \textbf{0.6756}  & \textbf{0.6460} \\ \hline
\end{tabular}
\end{center}
\caption{Performance of entity relation extraction on the FUNSD test data. We compare different pretrained language model used to encode entities and keep other modules of \textsc{\ourmodels{}} unchanged.}
\label{node_rep}
\end{table}

\subsection{Entity Representation}
While encoding semantic entities, we employ three different pretrained models\footnote{In our model, BERT and LayoutLM encode entities in each document as we concatenate words in all bounding boxes in the order from top left to bottom right as \citet{xu2020layoutlm}.} and comparison experiment results are listed in Table~\ref{node_rep}.
Results show that encoding entities with LayoutLM performs the best because it introduces layout information into its transformer encoder and has been pretrained on a large scale of VRDs compared with BERT.
Word2vec achieves much poorer performance than the other two due to the missing context-aware information inside semantic entities.

Table~\ref{full_exp_table} demonstrates that taking entity label embedding as part of the entity representation can improve the model performance.
From our analysis on the FUNSD data, we find that many entity relation links point from entities with label `Question' to entities with label `Answer' and almost no relations in-between answers or questions\footnote{In FUNSD training data, relations between answers or questions occupy only about 1\%.}.
Therefore, entity labels help the extraction model prune the unreasonable relations and enrich the entity representation.
The gap between models with gold labels and auto labels is about 10\% F1, which indicates the room for improvement is still large if we can obtain better auto entity labels.

\subsection{Document Encoder}

We investigate three popular encoders, including transformer, BiLSTM and GCN to encode VRDs. 
Experiment results in Table~\ref{encoder} show GCN encoder performs much better than the other two in our task.
The GCN we applied can be seen as an improvement of self-attention mechanism due to it introduces layout information into the document encoder as Formula~\ref{equation_h} shows.
Such layout features indicate the positional relation between entities according to x-axis or y-axis coordinates. 
They help document encoder to copy more information from adjacent entities which are more relevant to current entity, while transformer updates entity representation according to the textual information alone.
BiLSTM encodes the entities in documents in the plain sequential order. 
However, the sequential order is not suitable for the VRD understanding, for example, many key-value pairs in tables are in top and down order.

\begin{table}
\begin{center}
\small
\begin{tabular}{ l | c   c  c  }
\hline
 & \textbf{P} & \textbf{R} & \textbf{F1}  \\ \hline
\hline \textsc{Transformer} &   0.6022  & 0.6240 &   0.6129 \\
\hline \textsc{BiLSTM} &   0.6110  &  0.6460  &    0.6280    \\
\hline \textsc{GCN} & \textbf{0.6189} & \textbf{0.6756} & \textbf{0.6460} \\  \hline
\end{tabular}
\end{center}
\caption{Performance of entity relation extraction on the FUNSD test data. We compare different document encoder and other modules of \textsc{\ourmodels{}} remains unchanged.}
\label{encoder}
\end{table}

\begin{table}
\begin{center}
\small
\begin{tabular}{p{2.6cm}|p{0.95cm}<{\centering} p{0.95cm}<{\centering} p{1.05cm}<{\centering}}
\hline
 & \textbf{P} & \textbf{R} & \textbf{F1}  \\ \hline
\hline\textsc{MLP + multi} & 0.5317  &0.4803 & 0.5047   \\
\hline\textsc{MLP + single} & 0.3041 & 0.3082 &  0.3062  \\
\hline\textsc{Biaffine + multi} & \textbf{0.6470}  &  0.3618  &  0.4641    \\
\hline\textsc{Biaffine + single} & 0.6189  &  \textbf{0.6756}  & \textbf{0.6460}    \\\hline
\end{tabular}
\end{center}
\caption{Performance of \textsc{\ourmodels{}} on the FUNSD test data. Different relation scorers with different loss functions are used under different constraints.}
\label{sigle_multi_head}
\end{table}

\subsection{Relation Decoder}
Decoding relation links between entities with or without the single-head constraint leads to a large performance gap as Table~\ref{sigle_multi_head} shows.
Using different relation scorer, the trend between these two decoders is contrary.
MLP scorer performs poorer than biaffine scorer under single-head constraint.
This is because biaffine scorer is more suitable for single-head constraint which has been proved by \citet{dozat2016deep}.
Without such constraint, the task is similar to SRL, and more SRL previous works prefer to the MLP scorer.

Our error analysis finds that biaffine scorer without single-head constraint leads to the model prefers to predict multi-head links for entities, which is not consistent with the data distribution.

Due to the big gap between the biaffine scorer with single-head constraint and the MLP scorer with multi-head constraint, we finally choose the biaffine scorer and add the single-head constraint in our experiments.

\begin{table}
\begin{center}
\small
\begin{tabular}{ l | c c c }
\hline 
& \textbf{P} & \textbf{R} & \textbf{F1}  \\
\hline \hline
\textsc{\ourmodels{}} &0.6189  &  0.6756  &  0.6460   \\  
\hline \ \textsc{+ MTL} &  \textbf{0.6374}  &  0.6727  &  0.6546  \\
\hline \ \textsc{+ Aug} &  0.6315    & \textbf{0.6851}  &    0.6572  \\  
\hline \ \ \ \textsc{+ MTL} &  0.6368   & 0.6842 & \textbf{0.6596}  \\
\hline \ \ \ \textsc{+ Gold label} &  0.7402  & 0.7777 & 0.7585  \\
\hline

\end{tabular}
\end{center}
\caption{Performance of entity relation extraction on the FUNSD test data. We train our \textsc{\ourmodels{}} with two training strategies: data augmentation and multi-task learning.}
\label{trainstrategy}
\end{table}

\subsection{Training Strategies}
To get better performance in entity relation extraction task, we apply two training strategies.
Firstly, we take entity labeling and relation extraction tasks as multi-task learning (MTL) \cite{mtl_thero, pos_dep} and these two tasks share the pretrained language model in entity encoder and fine-tune the sharing parameter together during training.
Relation extraction task can improve by about 0.86\% F1 while labeling model performance drops a little from Table~\ref{trainstrategy}.
Performance improvement demonstrates that MTL is highly effective on alleviating error propagation from entity labeling task.


Secondly, we try to augment our training data due to the small size of training documents in the FUNSD data \cite{augmentation_cv}.
We randomly drop some words in ratio 0.2 from word group of each entity to obtain more pseudo documents.
We combine these pseudo training data with the gold training data and keep the test data unchanged.
Models trained on the training data after augmentation improve performance by about 1.1\% F1 with auto label and 2.9\% F1 with gold label.
Improvement gap between the two indicates that relation extraction model is sensitive to the accuracy of entity labels.

Combination of these two strategies performs the best under the auto entity label settings.

\begin{table}
\begin{center}
\small
\begin{tabular}{ l | c   c  c  }
\hline
& \textbf{P} & \textbf{R} & \textbf{F1}  \\ \hline
\hline
\multicolumn{4}{l}{\textbf{BERT}} \\ \hline
\ \  \textsc{English} & 0.7726 & 0.7806 & 0.7853 \\ \hline
\ \  \textsc{Chinese} & 0.7950 & 0.8320 & 0.8131 \\ \hline
\hline
\multicolumn{4}{l}{\textbf{LayoutLM}} \\ \hline
\ \  \textsc{English} & \textbf{0.8464} & \textbf{0.8602}  & \textbf{0.8533} \\  \hline
\end{tabular}
\end{center}
\caption{Performance of entity relation extraction on the customs data using \textsc{\ourmodels{}}. We compare different language models in different languages.}
\label{custom}
\end{table}

\subsection{Customs Data}
 
We apply our \textsc{\ourmodels{}} with best configuration to our collected customs data.
Due to documents in customs data may be in Chinese and English, pretrained Chinese or English language model cannot cover the words in documents by its vocabulary perfectly.
We conduct experiments with different pretrained models in different languages to study this problem deeply.

As Table~\ref{custom} shows, our proposed model works well on the customs data, whose scale is much larger than FUNSD.
Customs data contain more layout information, such as tables as Figure~\ref{custom_invoice_example} shows.
We observe that the Chinese BERT is better than the English BERT on our language mixed data.
We analyse their vocabularies and find Chinese vocabulary can cover more words in documents.
Even though Chinese BERT performs better, English LayoutLM still achieves the best results among three pretrained models.
This indicates encoding layout information into language model makes difference.

\section{Conclusion}
This paper focuses on the largely-unexplored entity relation extraction task in VRDs. We take advantages of previous works in semantic entity labeling and dependency parsing and propose our relation extraction model \textsc{\ourmodels{}}.
Our improved entity relation extraction model achieves 64.60\% F1 score on the FUNSD data, outperforming previous baseline by a large margin; and we employ two simple but effective training strategies to further improve the performance to 65.96\%, i.e., multi-task learning with entity labeling and data augmentation. In addition, We verify the effectiveness of our model on the real-world customs data with different layouts in the production setting. In the future, we plan to incorporate more visual features into the relation extraction model and also extend it into more domains and business scenarios.

\bibliography{anthology,custom}
\bibliographystyle{acl_natbib}

\end{document}